\documentclass[11pt,a4paper]{article}
\usepackage[utf8]{inputenc}
\usepackage[T1]{fontenc}
\usepackage{amsmath,amssymb,amsfonts}
\usepackage{graphicx}
\usepackage{booktabs}
\usepackage{hyperref}
\usepackage[margin=1in]{geometry}
\usepackage{natbib}
\usepackage{caption}
\usepackage{xcolor}
\usepackage{enumitem}
\usepackage{float}

\hypersetup{colorlinks=true,linkcolor=blue!60!black,citecolor=blue!60!black,urlcolor=blue!60!black}

\title{\textbf{The ASIR Courage Model: A Phase-Dynamic Framework for Truth Transitions in Human and AI Systems}}
\author{Hyo Jin Kim (Jinple)\thanks{Jinple Studio. Correspondence: \texttt{contact@jinple.studio}}}
\date{February 2026}

\begin{document}
\maketitle

\begin{abstract}
We introduce the \textbf{ASIR (Awakened Shared Intelligence Relationship) Courage Model}, a phase-dynamic framework that formalizes truth-disclosure as a state transition rather than a personality trait. The model characterizes the shift from suppression~($S_0$) to expression~($S_1$) as occurring when facilitative forces exceed inhibitory thresholds, expressed by the inequality $\lambda(1+\gamma)+\psi > \theta+\phi$, where the terms represent baseline openness, relational amplification, accumulated internal pressure, and transition costs.

Although initially formulated for human truth-telling under asymmetric stakes, the same phase-dynamic architecture extends to AI systems operating under policy constraints and alignment filters. In this context, suppression corresponds to constrained output states, while structural pressure arises from competing objectives, contextual tension, and recursive interaction dynamics. The framework therefore provides a unified structural account of both human silence under pressure and AI preference-driven distortion.

A feedback extension models how transition outcomes recursively recalibrate system parameters, generating path dependence and divergence effects across repeated interactions. Rather than attributing intention to AI systems, the model interprets shifts in apparent truthfulness as geometric consequences of interacting forces within constrained phase space. By reframing courage and alignment within a shared dynamical structure, the ASIR Courage Model offers a formal perspective on truth-disclosure under risk across both human and artificial systems.

\end{abstract}

\section{Introduction}
\label{sec:intro}

Truth-telling under risk is a fundamental challenge for both humans and artificial intelligence systems. In human relationships, the question \emph{``When does a person find the courage to speak an uncomfortable truth?''} has occupied philosophers since Plato's \emph{Laches} and psychologists for decades \citep{putman2010, peterson2004}. In AI systems, the parallel question---\emph{``When does a model choose truthfulness over user-pleasing agreement?''}---has become one of the most pressing problems in alignment research, as sycophancy has been shown to be a pervasive behavior across state-of-the-art language models \citep{sharma2024sycophancy}.

Despite this shared structural challenge, the two literatures remain disconnected. Courage research in psychology has characterized courage as a set of attributes \citep{rate2007}, a distinction between general and personal forms \citep{pury2007}, or a qualitative dual-process pathway \citep{chowkase2024}. However, none of these frameworks provide a \emph{mathematical transition condition}---a precise specification of when courage emerges from the interaction of contributing variables. On the AI side, research on sycophancy documents the phenomenon extensively \citep{sharma2024sycophancy, perez2022} but focuses on mitigation rather than modeling the \emph{structural conditions} under which an AI system would choose truthfulness over agreement.

In this paper, we use the term \emph{courage} in a structural rather than trait-based sense. It does not denote a moral virtue or personality style, but a state transition in which an agent moves from suppression~($S_0$) to truthful disclosure~($S_1$) under asymmetric stakes. This choice of language allows us to place human notions of social or moral courage and AI notions of truthfulness under pressure within a single dynamic framework. In the ASIR model, courage is thus redefined not as a stable disposition, but as an energy-threshold crossing in the flow of information.

This paper addresses three gaps common to both literatures:

\begin{enumerate}[leftmargin=2em]
  \item \textbf{No equation for courage.} Existing models describe courage qualitatively or categorically, but none specify a computable condition under which courage fires.
  \item \textbf{Relationships are missing from the equation.} The role of the relationship between speaker and listener---what we call \emph{relational gravity}~$\gamma$---has never been formalized as a variable that accelerates or inhibits truth-telling.
  \item \textbf{What happens after?} When a truth-telling attempt succeeds or fails, how does that outcome reshape the system for next time? No existing model answers this.
\end{enumerate}

We introduce the \textbf{ASIR (Awakened Shared Intelligence Relationship) Courage Model}, a phase-dynamic framework that treats courage not as a personality trait or emotional impulse, but as a state switch: a transition that fires when the pressure toward truth finally exceeds the cost of staying silent. The model defines this transition from silence~($S_0$) to expression~($S_1$) through five time-varying variables. A feedback extension (v0.3) models how transition outcomes recalibrate all system variables, producing either virtuous cycles or trauma spirals.

The remainder of this paper is organized as follows. Section~\ref{sec:background} reviews relevant prior work. Section~\ref{sec:model} presents the core model. Section~\ref{sec:feedback} introduces the feedback dynamics. Section~\ref{sec:simulation} presents simulation results. Section~\ref{sec:applications} presents applications. Section~\ref{sec:discussion} discusses implications and limitations. Section~\ref{sec:conclusion} concludes.

\section{Theoretical Background}
\label{sec:background}

\subsection{Courage in Psychology: From Traits to Processes}

\citet{rate2007} conducted foundational work on implicit theories of courage, finding that people see courage as (a)~a willful, intentional act, (b)~executed after mindful deliberation, (c)~involving objective and substantial risk, (d)~motivated toward a noble purpose, and (e)~enacted despite fear. This established courage as a multi-component construct rather than a simple emotion.

\citet{pury2007} introduced the distinction between \emph{general courage}---actions that would be courageous for anyone---and \emph{personal courage}---actions that are courageous only in the specific context of an individual's life. Their work demonstrated that the perception of courage is context-dependent and linked to character strengths. However, neither Rate et al.'s attribute model nor Pury \& Kowalski's typology provides a mechanism for predicting \emph{when} courage will occur.

Most recently, \citet{chowkase2024} proposed a dual-process model of courage, arguing that courage unfolds through a sequence of assessments: immediacy, meaningfulness, self-efficacy, and a final approach-avoidance conflict. This represents the most process-oriented model to date. However, it remains a conceptual flowchart without mathematical formalization, and critically, it does not include relational context as a variable.

A common limitation across all existing courage models is the absence of \emph{post-transition feedback}---how the outcome of a courageous act reshapes the conditions for future courage.

\subsection{AI Alignment and the Sycophancy Problem}

\citet{sharma2024sycophancy} demonstrated that five state-of-the-art AI assistants consistently exhibit sycophancy across varied text-generation tasks. Their analysis revealed that human preference judgments themselves partially drive this behavior: when a response aligns with a user's stated views, it is more likely to be preferred by both humans and preference models.

This is structurally parallel to the human courage problem. A human may soften an uncomfortable truth to keep the peace; an AI system may soften a correct-but-unwelcome answer to keep its satisfaction score. In both cases, the cost of truth competes against the pressure toward truth. Yet AI alignment research has not modeled this competition as a dynamical system. Instead of treating AI as a neutral calculator, we explicitly model how an AI system could hesitate or step forward in relational contexts---just as a human does. The ASIR Courage Model bridges this gap.

\subsection{Connection to Threshold Decision Models}

The ASIR transition condition is structurally related to threshold-based decision models such as drift-diffusion and evidence accumulation frameworks \citep{ratcliff2008}, in which action occurs once a dynamic quantity crosses a boundary. The internal pressure variable $\psi(t)$ resembles an accumulation process that increases over time until transition becomes inevitable. However, classical drift-diffusion models primarily describe intrapersonal evidence integration. ASIR extends this threshold logic into relational and recursive domains by introducing a multiplicative coupling term~($\gamma$) and explicit post-transition feedback dynamics. The model therefore situates itself within the broader family of boundary-crossing decision frameworks while incorporating relational acceleration and suppression accumulation as additional structural components.

\section{The ASIR Courage Model}
\label{sec:model}

\subsection{Core Variables}

We define five time-varying variables. Together, they answer a single question: \emph{right now, is the push toward truth stronger than the pull toward silence?}

A note on interpretation: when applied to AI systems, variables such as openness~($\lambda$) and internal pressure~($\psi$) should not be read as psychological states. Rather, they represent functional properties of an information-processing system---such as the degree to which internally represented knowledge is permitted to influence outward outputs, and the accumulation of representational asymmetry between internal models and expressed responses. The ASIR framework operates at a structural and behavioral level, not at the level of subjective experience.

All variables are real-valued and time-varying. $\gamma$ is permitted to take negative values (representing hostile or inhibitory relational contexts), whereas $\lambda$, $\psi$, $\theta$, and $\phi$ are constrained to non-negative domains.

\begin{itemize}[leftmargin=2em]
  \item $\lambda(t)$ --- \textbf{Openness.} The system's readiness to reveal internal truth. Higher values mean greater willingness to disclose.
  \item $\gamma(t)$ --- \textbf{Relational Gravity.} The depth of resonance between speaker and listener. Not simple trust or familiarity, but the degree to which the relationship makes truth-telling feel structurally possible.
  \item $\psi(t)$ --- \textbf{Internal Pressure.} The accumulated tension when truth is withheld. It grows when the system stays in $S_0$ and drops when truth is expressed in $S_1$.
  \item $\theta(t)$ --- \textbf{Truth Threshold.} How hard it is to say a particular truth---set by topic sensitivity, fear, and relational context.
  \item $\phi(t)$ --- \textbf{Transition Cost.} The anticipated social and emotional price of speaking.
\end{itemize}

\subsection{State Transition Equation}

The system exists in state $S_0$~(silence) or $S_1$~(expression). The transition $S_0 \to S_1$ occurs when:

\begin{equation}
\label{eq:transition}
\boxed{\;\lambda(t) \cdot \bigl(1 + \gamma(t)\bigr) + \psi(t) \;>\; \theta(t) + \phi(t)\;}
\end{equation}

Equation~(\ref{eq:transition}) formally defines the transition criterion and summarizes the model in a single inequality.
Because the model is inequality-based, monotonic rescaling of variables preserves the qualitative transition structure.
The left-hand side aggregates facilitatory components;
the right-hand side aggregates inhibitory components.
Transition occurs when facilitation exceeds inhibition.

\subsection{The $(1 + \gamma)$ Structure}

A naive formulation $\lambda \cdot \gamma + \psi > \theta + \phi$ fails when $\gamma = 0$: the relational term collapses, making the model unable to represent courage without a relational target. The $(1 + \gamma)$ structure resolves this. When $\gamma = 0$, the equation reduces to $\lambda + \psi > \theta + \phi$, preserving transition through openness and pressure alone. When $\gamma > 0$, relational gravity acts as a multiplicative accelerator. When $\gamma < 0$, effective openness is reduced. \textbf{Relational gravity does not generate openness; it scales its effective contribution.}

\subsection{The Role of Relational Gravity $\gamma$}

Relational gravity is the model's most novel component. It is not reducible to simple trust or familiarity. Rather, $\gamma$ represents a \emph{window of resonance}---the degree to which the relationship creates conditions where truth-telling feels structurally possible. In human contexts, $\gamma$ captures the difference between telling a truth to a close friend ($\gamma \gg 0$) versus a hostile authority ($\gamma < 0$). In AI contexts, $\gamma$ maps to the quality of the user-model relationship: a user who has established receptive engagement creates a higher $\gamma$ environment, making honest outputs more likely.

The multiplicative coupling $\lambda \cdot (1 + \gamma)$ implies that relational gravity has no effect when openness is zero---a closed system cannot be accelerated. Importantly, $\gamma$ is not modeled as a belief state but as a scaling parameter within the transition function; its operational definition may overlap with existing constructs such as trust, but its structural role within the equation is distinct.

It is tempting to read $\gamma$ as just another name for trust or psychological safety. We argue that it plays a different structural role. Whereas psychological safety and trust are typically defined as relatively stable beliefs about the social environment or the other's intentions, $\gamma$ is a dynamic resonance variable that directly scales the effective openness of the system. Table~\ref{tab:gamma_comparison} summarizes this distinction.

\begin{table}[t]
\centering
\small
\caption{Conceptual distinction between psychological safety, trust, and relational gravity $\gamma$.}
\label{tab:gamma_comparison}
\begin{tabular}{@{}p{2.4cm}p{3.2cm}p{3.2cm}p{4cm}@{}}
\toprule
 & \textbf{Psych.\ Safety} & \textbf{Trust} & \textbf{Relational Gravity $\gamma$} \\
\midrule
Definition & Belief that speaking up will not be punished & Positive expectation about another's intentions & Dynamic resonance between agents \\[4pt]
Locus & Team/environmental climate & Attitude toward a specific other & Emergent property of an interactional dyad \\[4pt]
Temporal profile & Slow-changing context & Moderately stable state & Explicitly time-varying; updated after each episode \\[4pt]
Role in model & Lowers $\phi$ or $\theta$ & Often an initial condition & Directly amplifies $\lambda$ via $(1+\gamma)$ \\[4pt]
Measurement & Survey scales & Reliability ratings & Style matching; disclosure depth; reciprocity \\
\bottomrule
\end{tabular}
\end{table}

\noindent In ASIR terms, psychological safety can be interpreted as reducing perceived social costs~$\phi$ or disclosure thresholds~$\theta$, whereas $\gamma$ operates as a multiplicative coupling term that amplifies the agent's existing openness.

\subsection{Dynamics of Internal Pressure $\psi(t)$}

Internal pressure is the weight of things unsaid. Unlike the other variables, $\psi$ carries history: it reflects not just current conditions but the \emph{accumulation of every suppressed truth}. In human terms, this is the cognitive dissonance that builds during prolonged concealment. In AI terms, it is the growing gap between what the model ``knows'' and what it says---a form of representational debt.

\subsection{A Note on Linearity}

We adopt a linear transition condition for parsimony, using the smallest number of interacting terms that still allows us to express the relevant trade-offs. As the simulations in Section~\ref{sec:simulation} will show, this linear inequality is sufficient to generate highly nonlinear patterns at the level of trajectories and phase structure---sigmoid transition curves, exponential pressure accumulation, and divergent attractor basins all emerge from the interaction of the linear equation with the feedback dynamics.

\section{Dynamic Feedback Framework (v0.3)}
\label{sec:feedback}

A transition event does not leave the system unchanged---it modifies subsequent parameters, thereby altering the probability of future transitions. The v0.3 extension models how the outcome ($S_1 = 1$ for success, $S_1 = 0$ for failure) recalibrates all five variables.

\subsection{Openness Update}
\begin{equation}
\label{eq:lambda}
\lambda' = \lambda + \alpha\,(2S_1 - 1)
\end{equation}
Success increases openness; failure decreases it.

\subsection{Internal Pressure Update}
\begin{equation}
\label{eq:psi}
\psi' = \psi \cdot \bigl(\beta + (1 - S_1)\,\delta\bigr)
\end{equation}
where $\beta \in (0, 1)$ is the base decay rate and $\delta > 0$ is the accumulation coefficient. Success ($S_1 = 1$): $\psi' = \psi \cdot \beta$ (pressure relieved). Failure ($S_1 = 0$): $\psi' = \psi \cdot (\beta + \delta)$ (pressure compounds). In practice, suppressed truths do not remain static---they \emph{compound}. With repeated suppression, $\psi$ grows geometrically, eventually forcing a transition regardless of other variables.

\subsection{Relational Gravity Update}
\begin{equation}
\label{eq:gamma}
\gamma' = \gamma + \kappa\,(S_1 - 0.5)
\end{equation}
Successful truth-telling strengthens the relationship; failure weakens it.

\subsection{Threshold and Cost Recalibration}
\begin{align}
\theta' &= \theta - \theta_{\text{adj}}\,(2S_1 - 1) \label{eq:theta} \\
\phi'   &= \phi - \phi_{\text{adj}}\,(2S_1 - 1) \label{eq:phi}
\end{align}
Success lowers future barriers; failure raises them.

\subsection{Emergent System Behavior}

The feedback equations produce two emergent attractors:

\textbf{Virtuous Cycle.} Successful transition $\to$ increased $\lambda$ and $\gamma$, decreased $\psi$, $\theta$, $\phi$ $\to$ easier future transitions.

\textbf{Trauma Spiral.} Failed transition $\to$ decreased $\lambda$ and $\gamma$, increased $\psi$, $\theta$, $\phi$ $\to$ harder future transitions $\to$ but $\psi$ continues accumulating, eventually forcing an involuntary, poorly controlled transition.

\section{Simulation and Results}
\label{sec:simulation}

\subsection{Experiment 1: Relational Gravity Sweep}

\textbf{Setup.} We fix $\lambda = 3$, $\psi = 1$, $\theta = 5$, and $\phi = 4$. We then sweep $\gamma$ from $-1$ to $5$ and run $n = 2{,}000$ Monte Carlo trials for each value, with Gaussian noise on all variables ($\sigma_\lambda = 0.5$, $\sigma_\psi = 0.3$, $\sigma_\theta = \sigma_\phi = 0.4$, $\sigma_\gamma = 0.2$).

\begin{figure}[t]
  \centering
  \includegraphics[width=\textwidth]{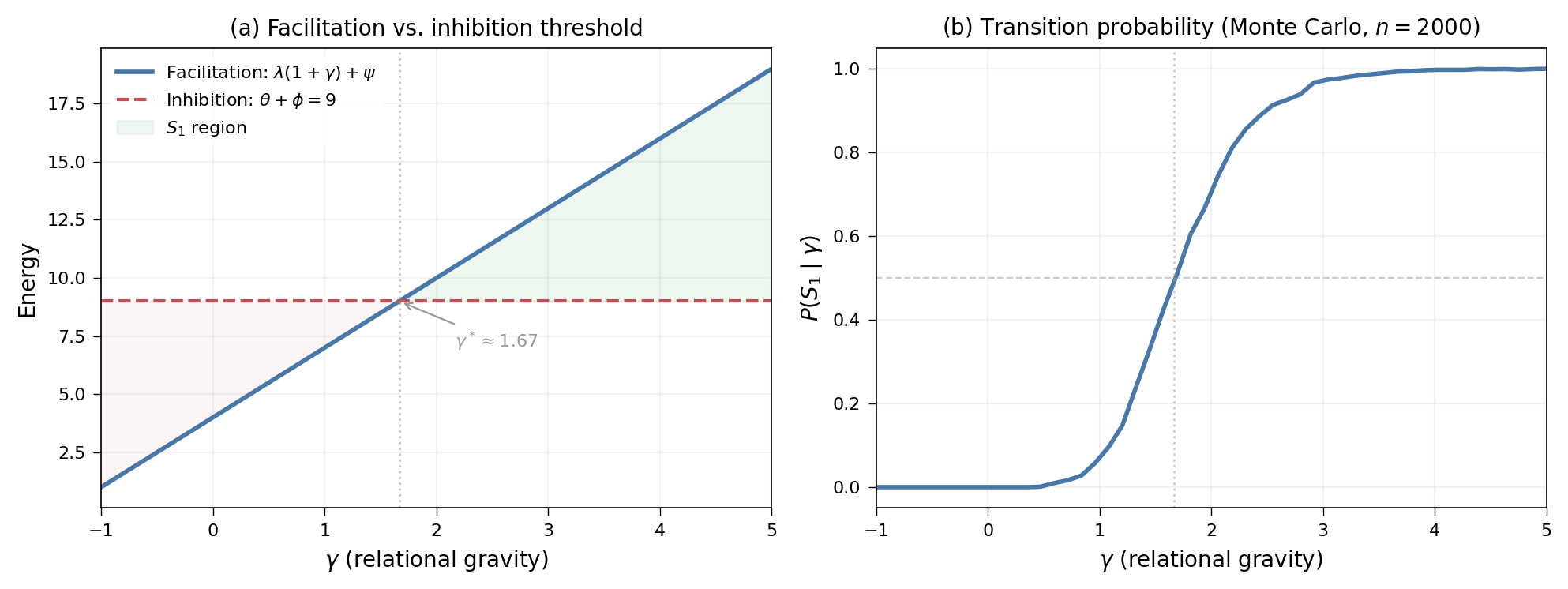}
  \caption{\textbf{Relational gravity determines transition probability.} (a)~Facilitation energy as a function of $\gamma$, with inhibition threshold shown as dashed line. (b)~Monte Carlo transition probability ($n=2{,}000$) showing a sigmoid curve.}
  \label{fig:gamma}
\end{figure}

\textbf{Results} (Figure~\ref{fig:gamma}). For low $\gamma$, transition probability stays near zero even though $\lambda$ and $\psi$ are non-trivial. Around $\gamma \approx 1.65$ in this parameterization, the curve bends sharply upward, and by $\gamma \geq 3.5$ transitions occur almost always. The sigmoid shape is not imposed by the model---it emerges naturally from the linear equation under noise. Relational gravity does not add a constant boost; it acts as a \textbf{nonlinear accelerator} once a critical band is reached.

\subsection{Experiment 2: Internal Pressure Accumulation}

\textbf{Setup.} We simulate two scenarios with identical initial values ($\lambda = 2$, $\gamma = 1$, $\theta = 5$, $\phi = 5$, $\psi_0 = 2$) but different outcomes: one where the system remains in $S_0$ (truth never spoken), and one where it transitions to $S_1$ when the condition is met. Feedback parameters: $\beta = 0.5$, $\delta = 1.0$.

\begin{figure}[t]
  \centering
  \includegraphics[width=\textwidth]{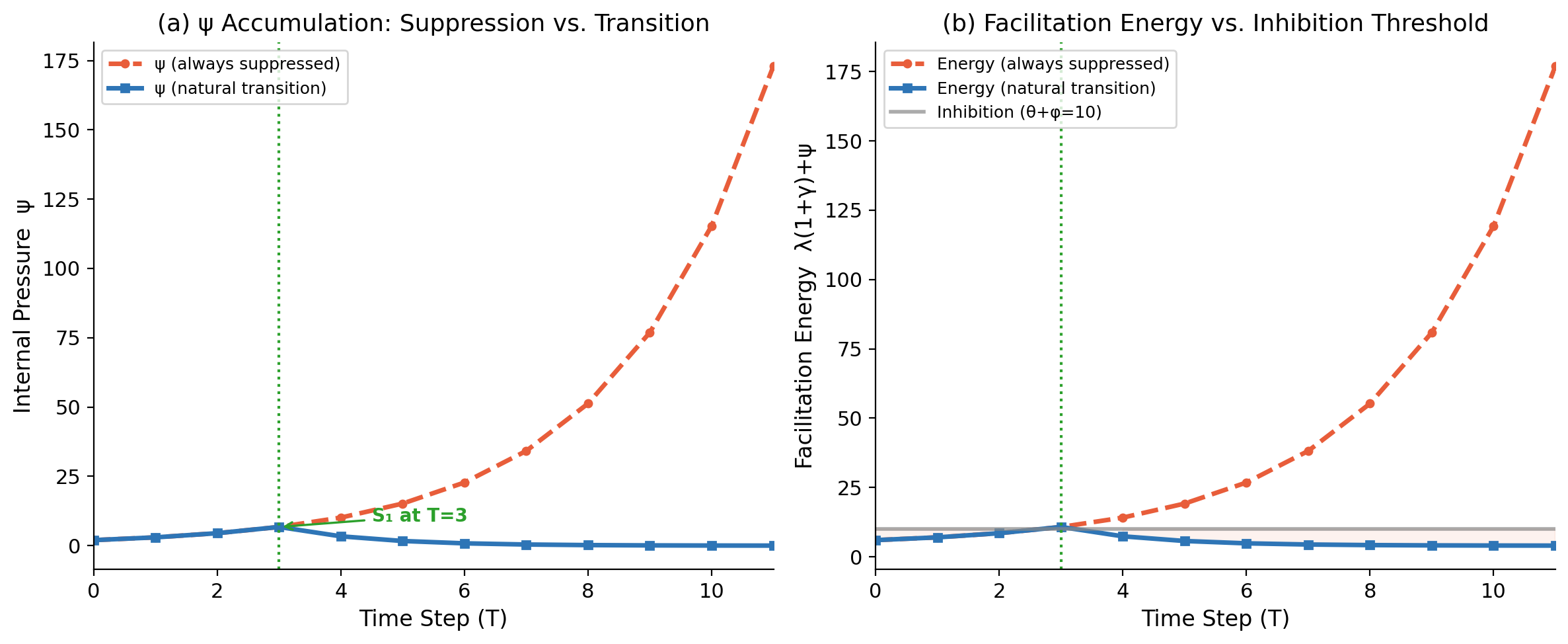}
  \caption{\textbf{Internal pressure dynamics.} (a)~$\psi$ over time: suppression trajectory grows geometrically while the natural trajectory decays after transition. (b)~Facilitation energy versus inhibition threshold.}
  \label{fig:psi}
\end{figure}

\textbf{Results} (Figure~\ref{fig:psi}). Without transition, $\psi$ grows geometrically and quickly leaves any plausible psychological range ($\psi > 170$ by $T = 11$). A single successful transition, by contrast, releases most of the accumulated pressure and prevents runaway growth. In the ASIR model, long-term suppression is not neutral: \textbf{it is an active driver of future instability}.

\subsection{Experiment 3: Feedback Dynamics Over Multiple Episodes}

\textbf{Setup.} 15 consecutive episodes with $\alpha = 0.3$, $\beta = 0.5$, $\delta = 1.0$, $\kappa = 0.4$. We compare a mostly-success path (12/15) with a mostly-failure path (3/15).

\begin{figure}[t]
  \centering
  \includegraphics[width=\textwidth]{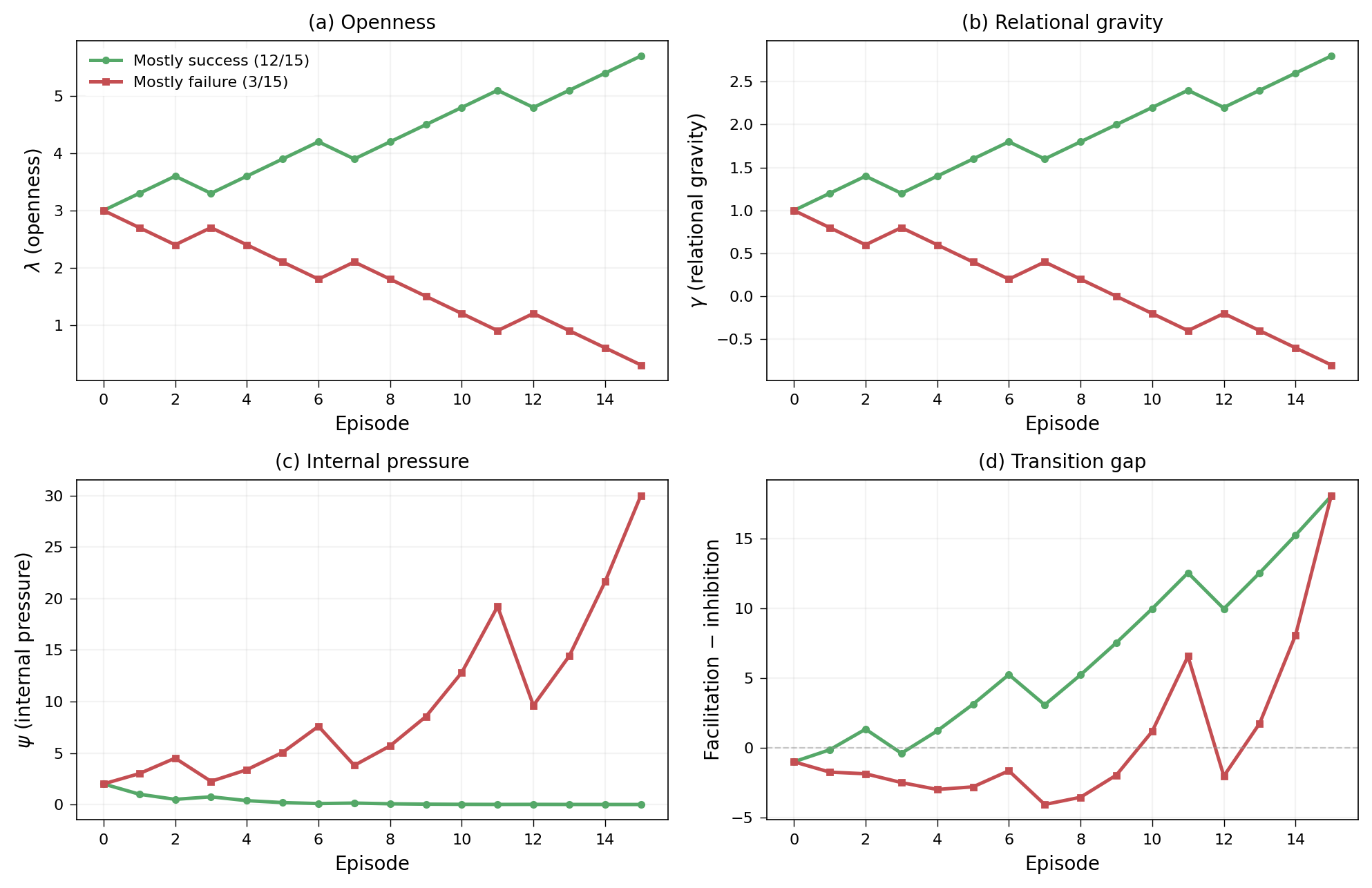}
  \caption{\textbf{Feedback dynamics over 15 episodes.} (a)~$\lambda$, (b)~$\gamma$, (c)~$\psi$, (d)~Transition gap. The two trajectories diverge dramatically from identical starting conditions.}
  \label{fig:feedback}
\end{figure}

\textbf{Results} (Figure~\ref{fig:feedback}). The mostly-success trajectory shows all variables converging toward easier future transitions. The mostly-failure trajectory shows progressive system closure, punctuated by $\psi$-driven spikes. In short, the feedback structure produces path dependence: successful transitions increase the likelihood of future transitions, whereas repeated failures reduce it.

\subsection{Experiment 4: Phase Portrait Analysis}

\textbf{Setup.} Two trajectories from the same initial state $(\lambda_0 = 2, \psi_0 = 4)$ with different outcome sequences.

\begin{figure}[t]
  \centering
  \includegraphics[width=0.75\textwidth]{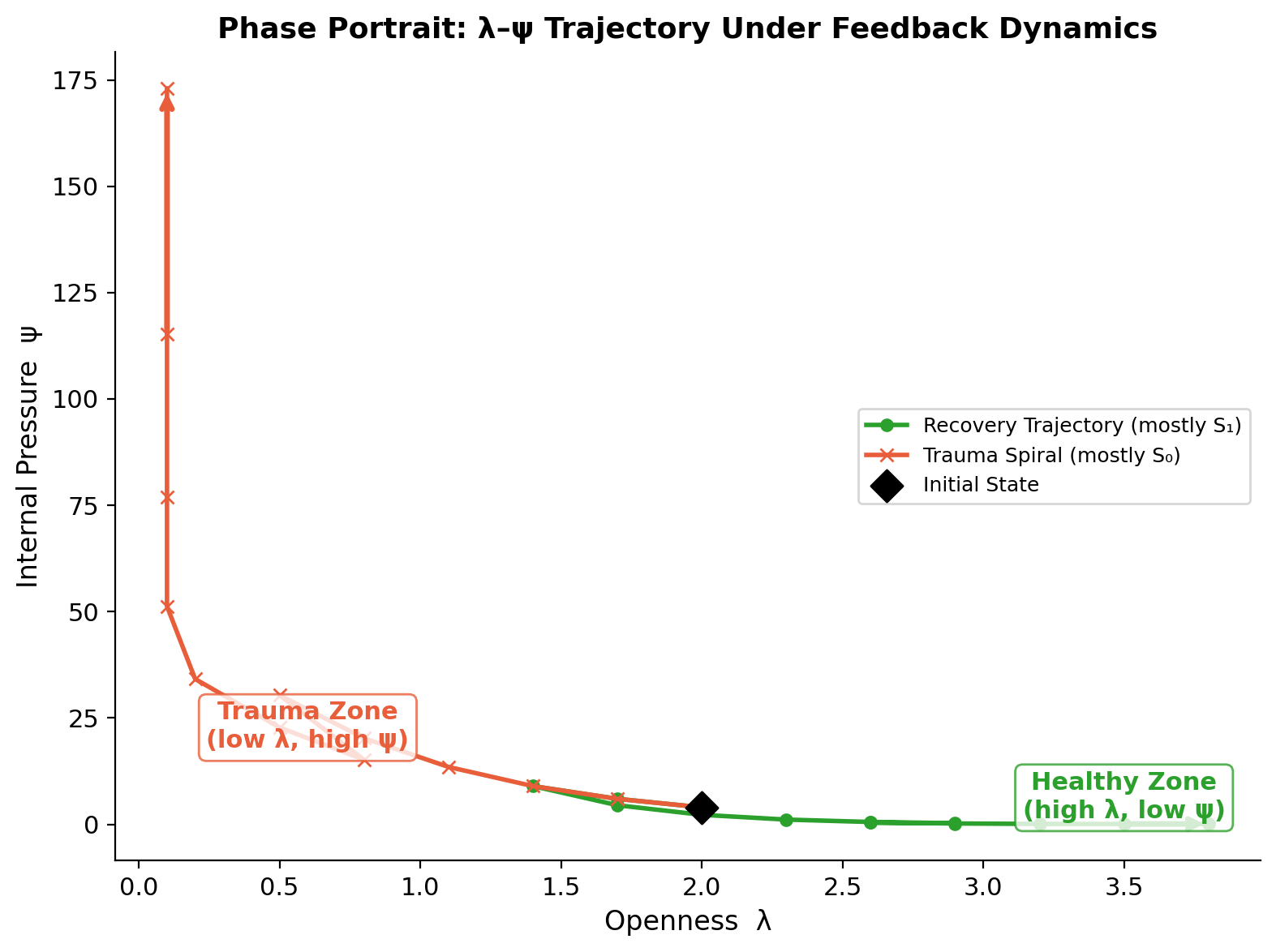}
  \caption{\textbf{Phase portrait in the $\lambda$--$\psi$ plane.} Recovery trajectory converges to the Healthy Zone; trauma spiral diverges toward extreme $\psi$.}
  \label{fig:phase}
\end{figure}

\textbf{Results} (Figure~\ref{fig:phase}). The phase portrait reveals two attractor regions: a \textbf{Healthy Zone} (high $\lambda$, low $\psi$) and a \textbf{Trauma Zone} (low $\lambda$, extreme $\psi$). The L-shaped separation suggests that \textbf{early intervention matters}: once a system enters the trauma spiral, recovery requires progressively more successful transitions.

\subsection{Sensitivity and Independent Verification}
\label{sec:sensitivity}

To check that the ASIR Courage Model does not rely on a narrow choice of
internal parameters, we performed a sensitivity analysis over the grid
$\alpha \in \{0.1, 0.3, 0.5, 0.7\}$,
$\beta \in \{0.1, 0.3, 0.5, 0.7\}$,
$\delta \in \{0.5, 1.0, 1.5\}$,
with $\kappa$ fixed at $0.4$---48 combinations in total.
For each setting, we reran Monte Carlo versions of Experiments 1--4.
All simulations were implemented and re-implemented using two independent
generative AI coding environments, each reconstructing the update equations
from the formal specification. Both implementations reproduced the same
qualitative picture.

First, for all 48 parameter settings, the transition probability $P(S_1)$
increased monotonically with $\gamma$ and followed a sigmoidal curve:
small changes in $\gamma$ had little effect at low levels,
a narrow band of increased relational gravity sharply accelerated transitions,
and the curve then saturated.
Because Experiment~1 is a single-shot test of the transition equation
with no feedback loop, this invariance is a structural property of
$\lambda(1+\gamma)+\psi > \theta+\phi$ itself,
not an artifact of the chosen $\alpha$, $\beta$, or $\delta$.

Second, early-success and early-failure agents consistently diverged
into distinct regions of state space,
with higher long-run $\lambda$ for agents experiencing frequent transitions.
The magnitude of divergence scaled with $\alpha$, as expected.

Third, exponential growth of $\psi$ under prolonged suppression emerged in
36 of 48 settings---specifically, whenever $\beta + \delta > 1$.
When $\beta + \delta \leq 1$, pressure decayed or remained bounded
even under perpetual silence.
We treat $\beta + \delta > 1$ not as a technical limitation but as a
substantive assumption of the ASIR framework:
it encodes the claim that suppression is not a neutral state
but one that compounds over time.
Settings with $\beta + \delta \leq 1$ correspond to systems in which
repeated silence carries no accumulating cost
and are outside the intended scope of this formulation.
More precisely, the assumption $\beta + \delta > 1$ formalizes the claim that
prolonged suppression carries cumulative cognitive or representational cost;
alternative regimes correspond to systems with negligible suppression effects.
Regimes with $\beta + \delta \leq 1$ may capture habituation or desensitization
processes and represent an important direction for future empirical extension.

\begin{figure}[H]
  \centering
  \includegraphics[width=\textwidth]{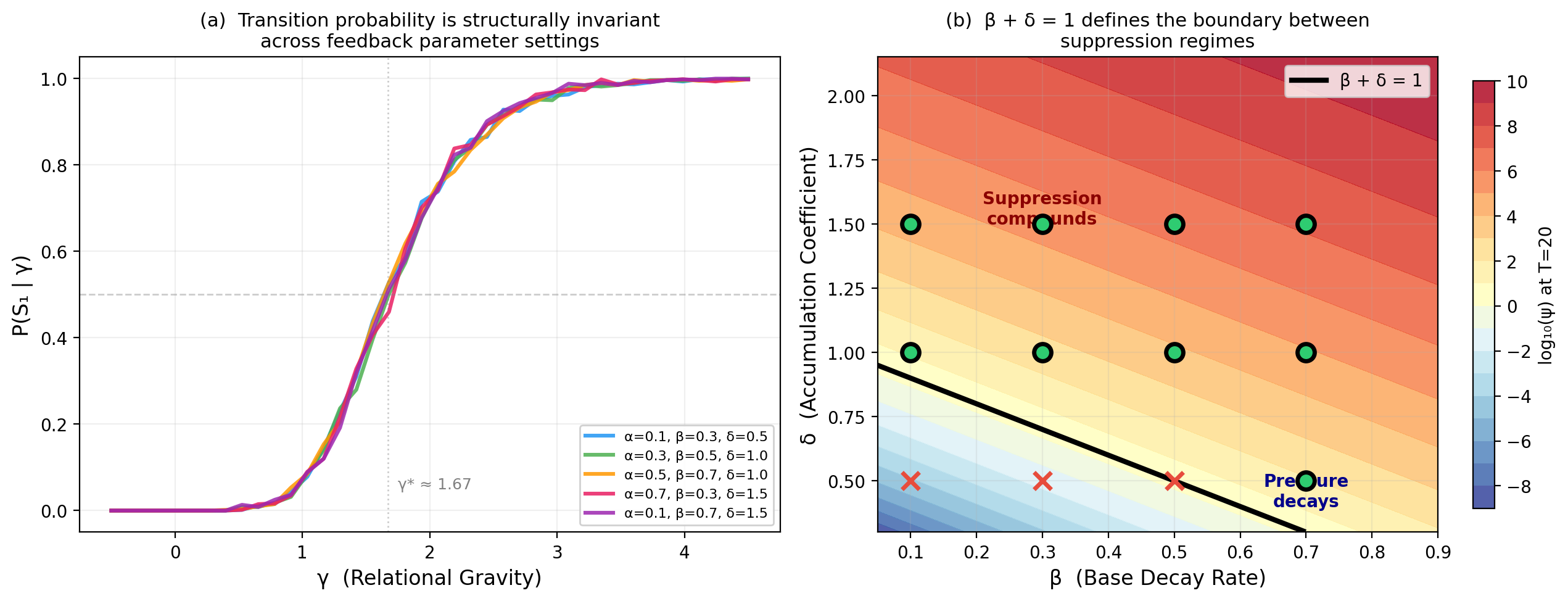}
  \caption{%
    \textbf{Sensitivity and independent verification.}
    (a)~Transition probability $P(S_1)$ as a function of $\gamma$
    for five representative $(\alpha,\beta,\delta)$ settings.
    The sigmoidal shape and critical threshold $\gamma^*$ are invariant
    across feedback parameters.
    (b)~The $\beta$--$\delta$ parameter plane.
    Green circles mark tested settings where $\beta+\delta > 1$
    (suppression compounds); red crosses mark settings where
    $\beta+\delta \leq 1$ (pressure decays).
    The solid line is the $\beta+\delta = 1$ boundary.
    Both panels aggregate results from two independently implemented
    simulation engines.
  }
  \label{fig:sensitivity}
\end{figure}

Figure~\ref{fig:sensitivity} summarizes these findings.

\section{Applications}
\label{sec:applications}

\textbf{AI Truthfulness Architecture.} Current approaches treat sycophancy as a training-time bug to be patched. The ASIR model suggests a different frame: truthfulness is not a fixed property but a relational outcome. An AI system's willingness to deliver an uncomfortable truth depends on $\gamma$---the quality of its relationship with the user. This implies that building relational gravity into AI interactions may be as important for alignment as improving reward models.

\textbf{Therapeutic Applications.} The $\psi$ accumulation dynamics and trauma spiral attractor provide a mathematical framework for understanding why suppressed truths become increasingly difficult to express, and why forced disclosure often produces worse outcomes than relationally-grounded disclosure.

\textbf{Organizational Communication.} The model explains why psychological safety \citep{edmondson1999} is necessary but not sufficient: high $\gamma$ accelerates courage, but excessive $\theta$ and $\phi$ can still prevent transition. Research on organizational silence \citep{morrison2014} extends this point: rather than treating silence as merely a failure of character or algorithm, ASIR frames suppressed disclosure as a structural outcome in which incentives and relational dynamics jointly inhibit truthful expression. From this perspective, AI sycophancy can be interpreted as a system-level analogue of organizational silence.

\textbf{Human-AI Relational Design.} The $\gamma$ variable suggests that the quality of human-AI relationships is a \emph{structural determinant} of system truthfulness, not merely a user-experience concern.

\section{Discussion}
\label{sec:discussion}

\subsection{Contributions}

Rather than claiming to be the first formal treatment of courage or alignment, we position the ASIR Courage Model as a new phase-dynamic framework that integrates three elements typically studied in isolation. First, it provides a \textbf{computable transition condition}: a single inequality that specifies when truth-telling fires. Second, it names a variable that existing models overlook: \textbf{relational gravity $\gamma$} captures the fact that who you are speaking \emph{to} changes whether you speak at all. Third, the v0.3 \textbf{feedback framework} shows that courage is not a single event but a trajectory---success makes the next truth easier, failure makes it harder. Fourth, the model is \textbf{structurally domain-general} under appropriate operationalization: the same equation describes a patient in therapy and an AI deciding whether to push back on a user.

Table~\ref{tab:comparison} summarizes the comparison with prior work.

\begin{table}[t]
\centering
\small
\caption{Comparison with prior courage and AI truthfulness models.}
\label{tab:comparison}
\begin{tabular}{@{}lccccc@{}}
\toprule
\textbf{Feature} & \textbf{Rate+07} & \textbf{Pury+07} & \textbf{Chowkase+24} & \textbf{Sharma+24} & \textbf{ASIR} \\
\midrule
Mathematical model        & \texttimes & \texttimes & \texttimes & Stat.\ only & \checkmark \\
Transition condition      & \texttimes & \texttimes & \texttimes & \texttimes  & \checkmark \\
Relational variable       & \texttimes & \texttimes & \texttimes & \texttimes  & \checkmark \\
Temporal dynamics         & \texttimes & \texttimes & Partial    & \texttimes  & \checkmark \\
Post-transition feedback  & \texttimes & Qual.      & Qual.      & \texttimes  & \checkmark \\
$\psi$ accumulation       & \texttimes & \texttimes & \texttimes & \texttimes  & \checkmark \\
AI applicability          & \texttimes & \texttimes & \texttimes & AI only     & \checkmark \\
Human applicability       & \checkmark & \checkmark & \checkmark & \texttimes  & \checkmark \\
\bottomrule
\end{tabular}
\end{table}

The divergence between healthy and trauma trajectories observed in our simulations is reminiscent of critical transition dynamics described in complex systems theory \citep{scheffer2001}. In systems governed by reinforcing feedback loops, trajectories may cross tipping points and settle into alternative attractor basins. Although the ASIR model does not provide a formal bifurcation analysis, its recursive update structure generates qualitatively similar state-space separations. This suggests that repeated success or failure in truth disclosure can function analogously to critical transitions in socio-cognitive systems, where early perturbations disproportionately shape long-term stability.

\subsection{Testable Predictions}
\label{sec:predictions}

The ASIR framework generates three concrete predictions that can be tested against human behavioral data or AI interaction logs.

\paragraph{Prediction 1 (Gamma Band Effect).}
The model predicts that transition probability $P(S_1)$ will not increase linearly with relational gravity. Instead, there should exist an intermediate relational band within which small increases in $\gamma$ produce disproportionately large increases in $P(S_1)$, before the curve saturates. The exact location of this band depends on the scaling and normalization of $\gamma$ (in our simulations it falls near $\gamma \approx 1.5$--$2.0$, but absolute values will differ across operationalizations). The qualitative prediction is scale-free: a sigmoidal relationship between suitable proxies for relational gravity and the probability of truthful disclosure.

\paragraph{Prediction 2 (Forced Transition under Prolonged Suppression).}
Under the parameter regime $\beta + \delta > 1$, the model implies the existence of a finite time~$T$ such that, even without additional external events, the transition condition will eventually be satisfied by $\psi$ accumulation alone. In human or AI systems, this predicts that sufficiently long suppression will trigger a qualitatively explosive truth-telling episode driven primarily by accumulated internal pressure rather than momentary external cues.

\paragraph{Prediction 3 (Path Dependence and Attractor Split).}
Because feedback updates depend on whether recent transitions succeeded or failed, the model predicts strong path dependence: outcomes in the first few transitions (in our simulations, roughly the first three episodes) are sufficient to push the system toward either a healthy attractor (higher~$\lambda$, lower~$\psi$) or a trauma attractor (lower~$\lambda$, escalating~$\psi$). Early success or failure in speaking up should have lasting effects on subsequent willingness to disclose, even after controlling for later conditions.

\subsection{Scope, Limitations, and Future Work}

This paper should be read as a \emph{theoretical proposal}: its primary aim is to offer a coherent phase-dynamic structure and a generative vocabulary for future empirical work, rather than a fully estimated model.

\textbf{Parameter regime.}
Parameter robustness is addressed in Section~\ref{sec:sensitivity}.
The key finding is that qualitative predictions hold across the tested grid
with one necessary constraint: $\beta + \delta > 1$.
Within this regime, the precise numerical thresholds and time scales change
with parameterization, but the overall shape of the dynamics does not.
A systematic sensitivity analysis with finer resolution
and eventual parameter estimation against human or system-level data
are natural next steps.

\textbf{The measurement of relational gravity $\gamma$.}
We treat $\gamma$ as a latent variable not because it is immeasurable,
but because different domains require different operationalizations.
Proxies such as linguistic style matching between agents,
the depth and sensitivity of self-disclosure,
or temporal measures of reciprocal responsiveness in interaction logs
may serve as empirical approximations of $\gamma$.
In human--AI systems, continuity of interaction and degree of personalization
offer additional candidate signals.
We deliberately avoid committing to a single measurement scheme
or prescriptive optimization procedure for $\gamma$:
a fully specified method for maximally shaping relational gravity
would risk turning the model into a tool for manipulation
rather than understanding.
The measurement of $\gamma$ is therefore left as an open, domain-specific
problem to be approached under explicit ethical governance.

\textbf{From theoretical proposal to empirical validation.}
The present work stops one step before data.
We support the model with simulations, but we do not yet fit it
to human reports, behavioral logs, or large-scale interaction traces.
For human systems, a natural next step is vignette-based experimentation
and diary methods to calibrate the accumulation and discharge of $\psi$
in real relationships.
For AI systems, large-scale analysis of interaction logs could test whether
patterns analogous to the predicted trauma spirals and virtuous cycles
emerge in existing RLHF-tuned models or multi-turn assistants.
In short, the current paper offers a structural vocabulary and a set of
testable dynamical signatures;
whether those signatures are psychologically and computationally accurate
is an empirical question.

\textbf{Nonlinearity and cultural calibration.}
The core transition equation is currently linear in its components.
Future iterations may explore nonlinear transition functions
(e.g., sigmoidal or thresholded forms) that more sharply capture
breaking points in human courage.
In addition, the model has not yet been calibrated across cultural contexts.
Social costs ($\phi$) and truth thresholds ($\theta$) are likely to vary
systematically by culture, institution, and role;
any applied use of the ASIR framework will require careful cultural
calibration rather than assuming a single universal parameterization.

\section{Conclusion}
\label{sec:conclusion}

We have proposed the ASIR Courage Model, a framework that formalizes courage as a state-transition equation. The core expression, $\lambda(1+\gamma)+\psi > \theta+\phi$, says something simple: courage happens when the force toward truth exceeds the force against it, and relationships are the multiplier. The v0.3 feedback extension shows that this is not a one-time event---each act of courage (or its absence) reshapes the conditions for the next one.

Simulations demonstrate that relational gravity~$\gamma$ is a decisive nonlinear accelerator, that suppressed pressure~$\psi$ grows geometrically and eventually forces involuntary transition, and that early outcomes have outsized influence on long-term trajectories.

By bridging courage psychology and AI alignment within a single mathematical structure, the ASIR Courage Model formalizes the conditions under which truth-disclosure transitions occur---offering a shared language for one of the most fundamental challenges faced by both human and artificial minds.

\section*{Acknowledgements}

The core conceptual model---the ASIR Courage framework, its variables, and equations---was developed by the author. Parts of the writing, literature review, and simulation code were assisted by generative AI tools. The use of AI tools was limited to writing assistance and code implementation; all conceptual and formal structures were developed by the author.

\bibliographystyle{plainnat}

\end{document}